\documentclass[letterpaper,10pt,conference,final]{ieeeconf}

\IEEEoverridecommandlockouts
\usepackage{graphics} 
\usepackage{mathptmx} 
\usepackage{amsmath} 
\usepackage{amssymb} 
\usepackage{cite}
\usepackage{enumerate}
\usepackage{xcolor}
\usepackage{graphicx}
\usepackage{amsfonts}
\usepackage{mathtools}
\usepackage[mathcal]{euscript}
\usepackage{caption}
\usepackage[hidelinks]{hyperref}
\usepackage[T1]{fontenc} 
\usepackage[hyphenbreaks]{breakurl} 
\usepackage{subcaption}
\usepackage[ruled,vlined]{algorithm2e}

\DeclareCaptionLabelSeparator{periodspace}{.\quad}
\usepackage{amsthm}

\theoremstyle{definition} 

\title{\LARGE \bf
	Vision-Based Distributed Formation Control of  \\ Unmanned Aerial Vehicles
}

\author{Kaveh Fathian, Emily Doucette, J. Willard Curtis, Nicholas R. Gans
	\thanks{*This work was supported in part by the U.S. Air Force Research Laboratory under Grant FA8651-17-1-0001.}
	\thanks{K. Fathian and N. R. Gans are with the Department of Electrical Engineering, University of Texas at Dallas, Richardson, TX, 75080 USA.~  E-mail: {\tt\small \{kaveh.fathian, ngans\}@utdallas.edu}        }%
	\thanks{E. A. Doucette and J. W. Curtis are with the Air Force Research
	Laboratory, Munitions Directorate, Eglin AFB, FL, 32542, USA.~ E-mail: {\tt\small \{emily.doucette, jess.curtis\}@us.af.mil}			}.%
}

\def\bn{\mathbb N}

\def\br{\mathbb R}

\begin{document}
	
\maketitle
\thispagestyle{empty}
\pagestyle{empty}

\begin{abstract}
We present a novel control strategy for a team of unmanned aerial vehicles (UAVs) to autonomously achieve a desired formation using only visual feedback provided by the UAV's onboard cameras.
This effectively eliminates the need for global position measurements. The proposed pipeline is fully distributed and encompasses a collision avoidance
scheme. 
In our approach, each UAV extracts feature points from captured images 
and communicates their pixel coordinates and descriptors among its neighbors. These feature points are used in our novel pose estimation algorithm, QuEst, to localize the neighboring UAVs. Compared to existing methods, QuEst has better estimation accuracy and is robust to feature point degeneracies. 
We demonstrate the proposed pipeline in a high-fidelity simulation environment and show that UAVs can achieve a desired formation in a natural environment without any fiducial markers.
\end{abstract}

\section*{Supplementary Material}

Video of the simulations is available at 
{\color{blue} \href{https://youtu.be/AMXbX1Ezg8Q}{https://youtu.be/AMXbX1Ezg8Q}}, and the  code can be accessed at 
{\color{blue} \href{https://goo.gl/QH5qhw}{https://goo.gl/QH5qhw}}.

\section{Introduction}

In recent years, Unmanned Aerial Vehicles (UAVs) have persistently become more miniaturized and affordable, while their onboard computational and communication capabilities have advanced significantly.
Thanks to these technological trends and theoretical advances in distributed control, it is now possible to deploy UAVs to collaboratively map and monitor an environment  \cite{Cieslewski2018}, inspect infrastructures \cite{Ozaslan2017}, deliver goods \cite{Dorling2017}, or manipulate objects \cite{Baehnemann2017, Loianno2018}.
In these applications, the ability to bring the UAVs to a desired geometric shape is a fundamental building block upon which more sophisticated maneuvering and navigation policies can be constructed.

There exists a large body of work on formation control of autonomous vehicles \cite{Barca2013, Yan2013, Oh2015} that can be used to achieve a desired configuration. However, many methods rely on a centralized motion planning scheme or a global positioning/communication paradigm \cite{Michael2008, Aranda2015, Hyun2016,  Motoyama2017}.
Fully distributed formation control strategies \cite{Trinh2018, Fathian2016, Fathian2016a, Han2018}, on the other hand, do not have these requirements and compared to the centralized methods have better scalability, naturally parallelized computation, and resiliency to global positioning signal jamming or loss.

\begin{figure} [t!]
	\begin{center}
		\includegraphics[trim =10mm 15mm 10mm 10mm, clip, width=0.486\textwidth]{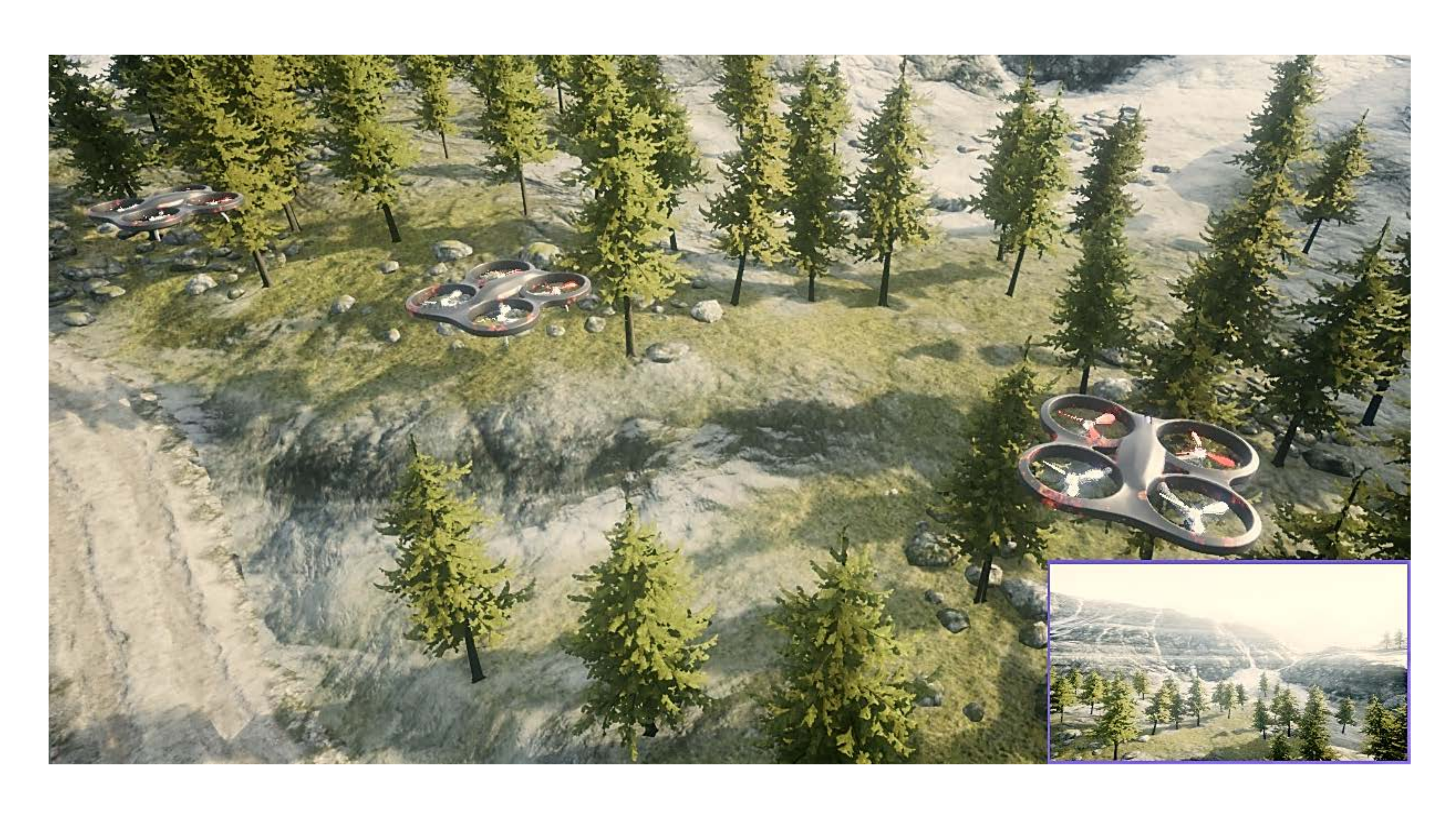}	
		\caption{A simulation of our distributed formation control strategy with 3 quadrotors in Unreal Engine 4 using Microsoft AirSim. Bottom right: Image provided by the onboard camera of the first quadrotor. Onboard images are used to estimate the relative pose between quadrotors, which is needed for formation control.}
		\label{fig:AirSim}
	\end{center}
\end{figure}

The main contribution of this abstract is to address two major challenging problems that are often not considered in the distributed formation control literature:  
1) How can UAVs achieve a desired formation without relaying on a centralized motion planning scheme to avoid collision?
2) How can UAVs localize their neighbors without using global position measurements?
To address the first question, we propose a control design that enjoys several properties which can be used for collision avoidance. 
To address the second questions, we propose to use the images captured by onboard UAV cameras in a novel camera pose estimation routine to estimate the relative position and orientation information needed for the formation control. 
The collision avoidance scheme that naturally arises from our control desing is what distinguishes our pipeline the most from similar work \cite{Montijano2016, Vemprala2018} that use visual feedback for localization.

\section{Methodology} \label{sec:Methodology}

In what follows, we present the formation control and camera pose estimation components of the proposed pipeline and show how they can be merged as a full system.
We assume that the UAVs are desired to operate at a constant altitude, all desired formations are planar, and the UAVs are equipped with low-level flight controllers that can regulate their speed to a desired value.

\subsection{Distributed Formation Control} 

To bring a team of $n$ UAVs to a desired planar formation, we define a linear velocity control vector $u_i \in \br^{2}$ for each UAV as
%
\begin{gather} \label{eq:HolonomCtrl}
u_i := \sum_{j \in \mathcal{N}_i}{A_{ij} \, q_j^{\,i} },
\end{gather}
%
where $i \in \{1,\, 2,\, \dots,\, n \}$ denotes the index of each UAV, $\mathcal{N}_i \subset \bn$ is the index set of neighboring vehicles for the $i$-th UAV, $q_j^{\,i} \in \br^{2}$ denotes the position of UAV $j$ in UAV $i$'s local coordinate frame, and $A_{ij} \in \br^{2\times 2}$ are constant control gain matrices that are provided to each UAV before the mission. 
At any instant of time, the low-level flight controller of each UAV regulates the UAV's linear velocity to the desired control vector $u_i$ given in \eqref{eq:HolonomCtrl}.

If the $A_{ij}$ matrices are chosen properly, the desired formation emerges from the interaction of all vehicles \cite{Lin2014, Lin2016}. To design these matrices, we propose a novel approach based on a semidefinite problem (SDP) formulation that not only ensures that a desired formation is achieved, but further has the following properties that are proven in our previous work \cite{Fathian2018, Fathian2018b}:
\begin{enumerate}
	\item The control is robust to perturbations, noise, and disturbances in the input.
	\item The control is robust to unmodeled dynamics and input saturations.
	\item Any positive scaling of the control vector does not affect the convergence of UAVs to the desired formation.
	\item If UAVs move in the desired direction perturbed by a rotation up to $\pm 90^\circ$, convergence to the desired formation remains guaranteed.
\end{enumerate}

We exploit property (4) to design a fully distributed collision avoidance strategy. In this strategy,  if moving along the control vector results in  collision, the UAVs rotate their control vectors to avoided collision. If the required rotation is above $90^\circ$, the UAVs stop until a feasible direction is available. Although gridlocks can occur and are unavoidable due to the distributed nature of this strategy, in our extensive simulations we have observed that if the UAVs are initially far apart they can often overcome gridlocks and converge to the desired formation.

If the sensing/communication graph among UAVs is time-varying, by adding addition constraints to the SDP problem one can ensure that the UAVs achieve the desired formation regardless of the switches in the sensing topology, as shown in our previous work \cite{Fathian2017}.

\subsection{Camera Relative Pose Estimation}

\begin{figure} [t!]
	\begin{center}
		\includegraphics[trim =45mm 73mm 53mm 70mm, clip, width=0.486\textwidth]{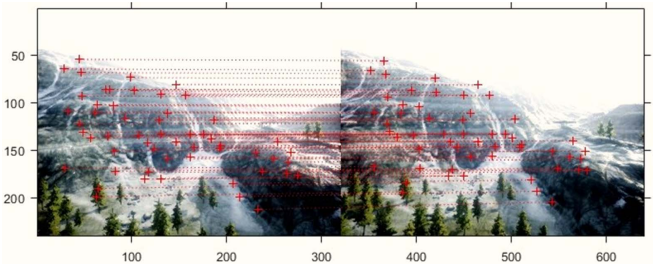}	
		\caption{Two $320\times 240$ images captured by UAVs' front facing cameras shown side-by-side. Image feature points are shown by $+$ signs and are matched between the images as illustrated by dotted lines.}
		\label{fig:MatchedPts}
	\end{center}
\end{figure}

Image feature points are edges, corners, etc., in an image, which can be identified and matched in images that were taken from a scene at two different locations. An example is shown in Fig.~\ref{fig:MatchedPts}, where SURF feature points \cite{Bay2006} indicated by $+$ signs are extracted from the onboard images of two UAVs and matched as illustrated with the dashed lines. The homogeneous coordinates of matched feature point pairs, $m,\, n \in \br^{3}$, are known from the images and must satisfy the rigid motion constraint
%
\begin{gather} \label{eq:RigidMotion}
u \, R \, m + t = v\, n,
\end{gather}
%
where $R \in \mathrm{SO}(3)$ and $t \in \br^{3}$ are respectively the unknown relative rotation matrix and the translation vector (i.e., pose) between the camera coordinate frames at each view, and $u,\, v \in \br$ are unknown depths of feature points in the corresponding coordinate frames. 
By using the quaternion representation of rotation matrix $R$ in \eqref{eq:RigidMotion}, our novel pose estimation algorithm, QuEst, can recover the relative pose from as low as 5 generic feature point pairs. On average, QuEst has shown up to 50\% increase in the estimation accuracy over state-of-the-art algorithms \cite{Fathian2018a}. 
Furthermore, random sample consensus (RANSAC) \cite{Fischler1981} can be incorporated to eliminate any potential mismatched feature points.

\subsection{Vision-Based Distributed Formation Control} 

In our proposed formation control pipeline, the relative position measurements required to compute the control vector \eqref{eq:HolonomCtrl} are estimated from the images captured by the onboard UAV cameras.
Feature points pixel coordinates and their descriptors are extracted from onboard images and communicated by UAVs to their neighbors.  
If the cameras have overlapping fields of view, corresponding feature points can be matched based on their descriptors and used in QuEst to estimate the relative pose. 
It is well-known that by solely using the images the estimated relative positions can only be recovered up to a common positive unknown scale factor. This, however, does not affect the convergence of the UAVs to the desired formation due to the property (3) in our proposed formation control strategy.

To investigate the required communication bandwidth of our proposed pipeline, we consider SURF image features \cite{Bay2006} and assume that 500 feature points are extracted in 50 ms from $320 \times 240$ pixel images. Each feature point consists of 2 single precision $x$-$y$ coordinates and a 64 element single precision descriptor vector, where each single precision number is represented by 32 bits. Hence, the communication bandwidth required for each neighboring UAV is about 5.2 KB/s, which is well within the range of current wireless communication technology.

\section{Simulation Results} \label{sec:Simulations}

\begin{figure*}[t!]
	\centering
	\begin{subfigure}[b]{0.5\textwidth}
		\includegraphics[trim = 10mm 35mm 10mm 35mm, clip, width=1.0\textwidth] {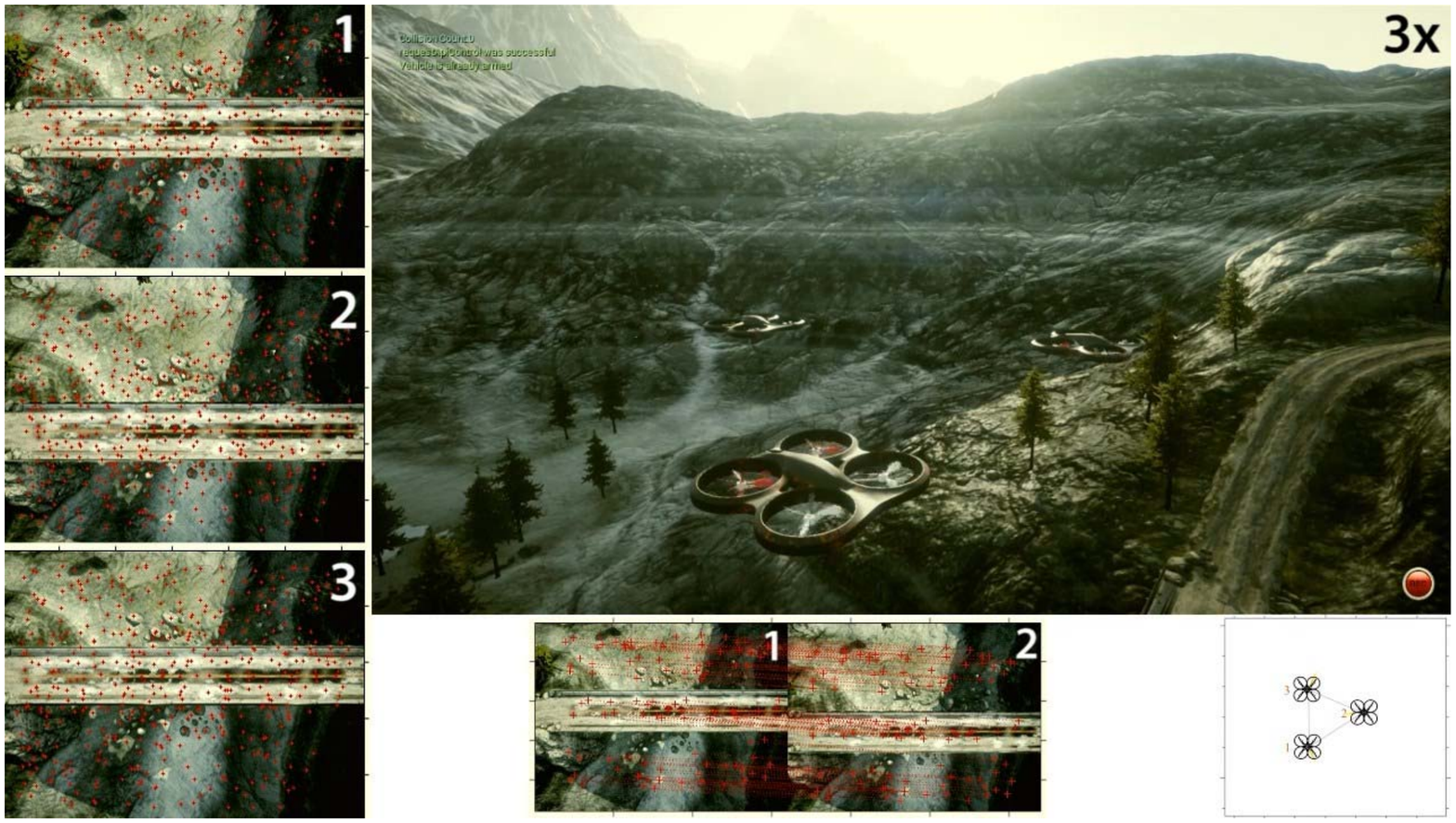}
		\caption{}
	\end{subfigure}%
	\begin{subfigure}[b]{0.5\textwidth}
		\includegraphics[trim = 10mm 35mm 10mm 35mm, clip, width=1.0\textwidth] {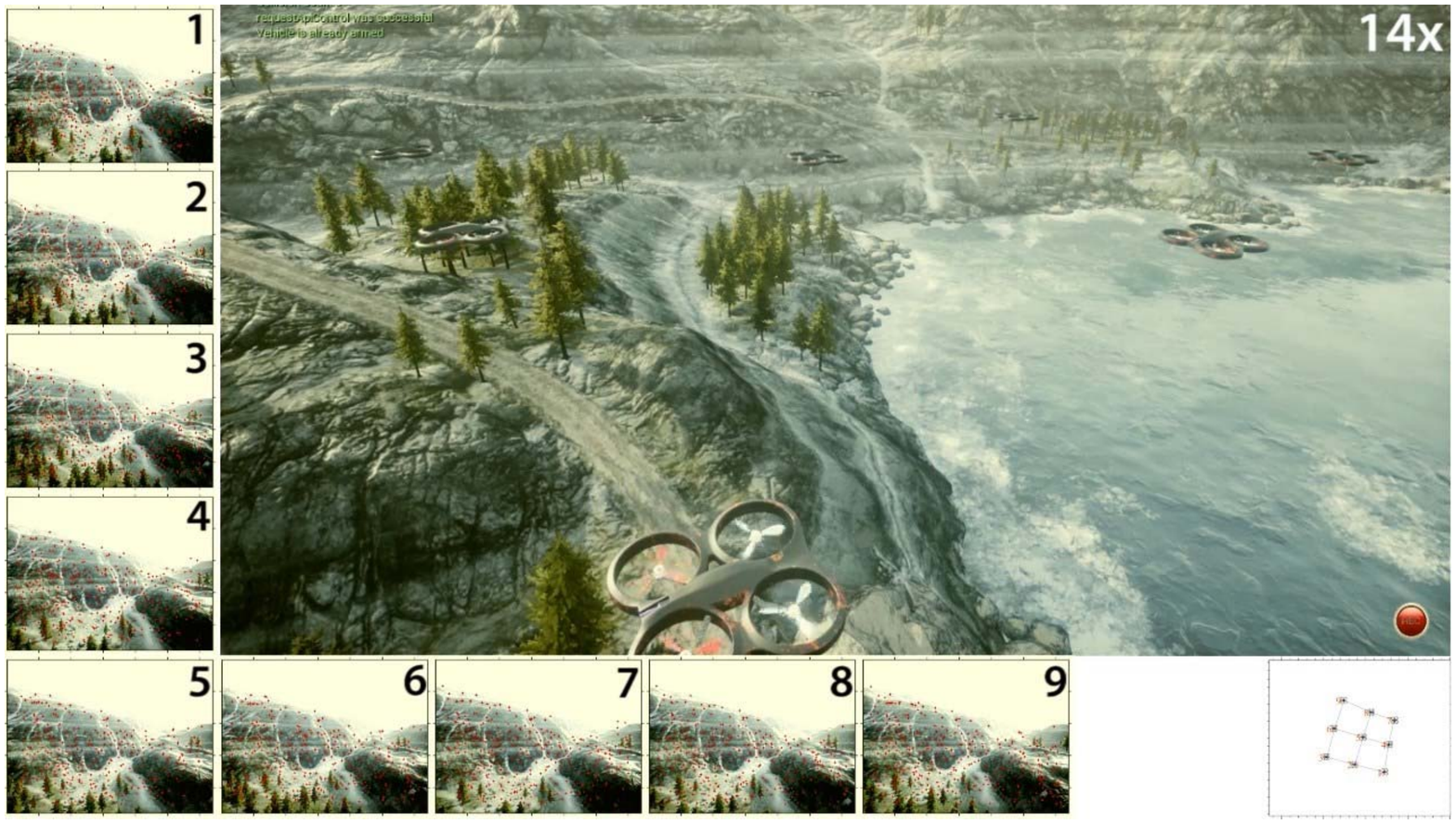}
		\caption{}
	\end{subfigure}%
	\caption{Snapshots of simulation videos for (a) 3 quadrotors with an equilateral triangle desired formation. (b) 9 quadrotors with a square grid desired formation.}
	\label{fig:Sims}	
\end{figure*}

We conduct several simulations for a formation of quadrotors using their front-facing or downward-facing cameras.
Our simulations verify that by using the proposed pipeline a team of quadrotors can achieve a desired formation without collision and by solely using their onboard images.
The simulations are performed in the high-fidelity Unreal Engine 4 environment using Microsoft AirSim \cite{Shah2017}.  AirSim is designed according to the software-in-the-loop (SITL) principle, in which the quadrotor flight controller cannot specify whether it is running under simulation or a real vehicle. The simulator includes PX4 SITL flight controller, which allows the simulation code to be directly imported to the commercially available quadrotor platforms. 
Links to simulation code and video are provided in the Supplementary Material section.

\begin{figure}[t]
	\centering
	\begin{subfigure}[b]{0.5\linewidth}
		\includegraphics[trim = 1mm 1mm 1mm 1mm, clip, width=0.97\textwidth] {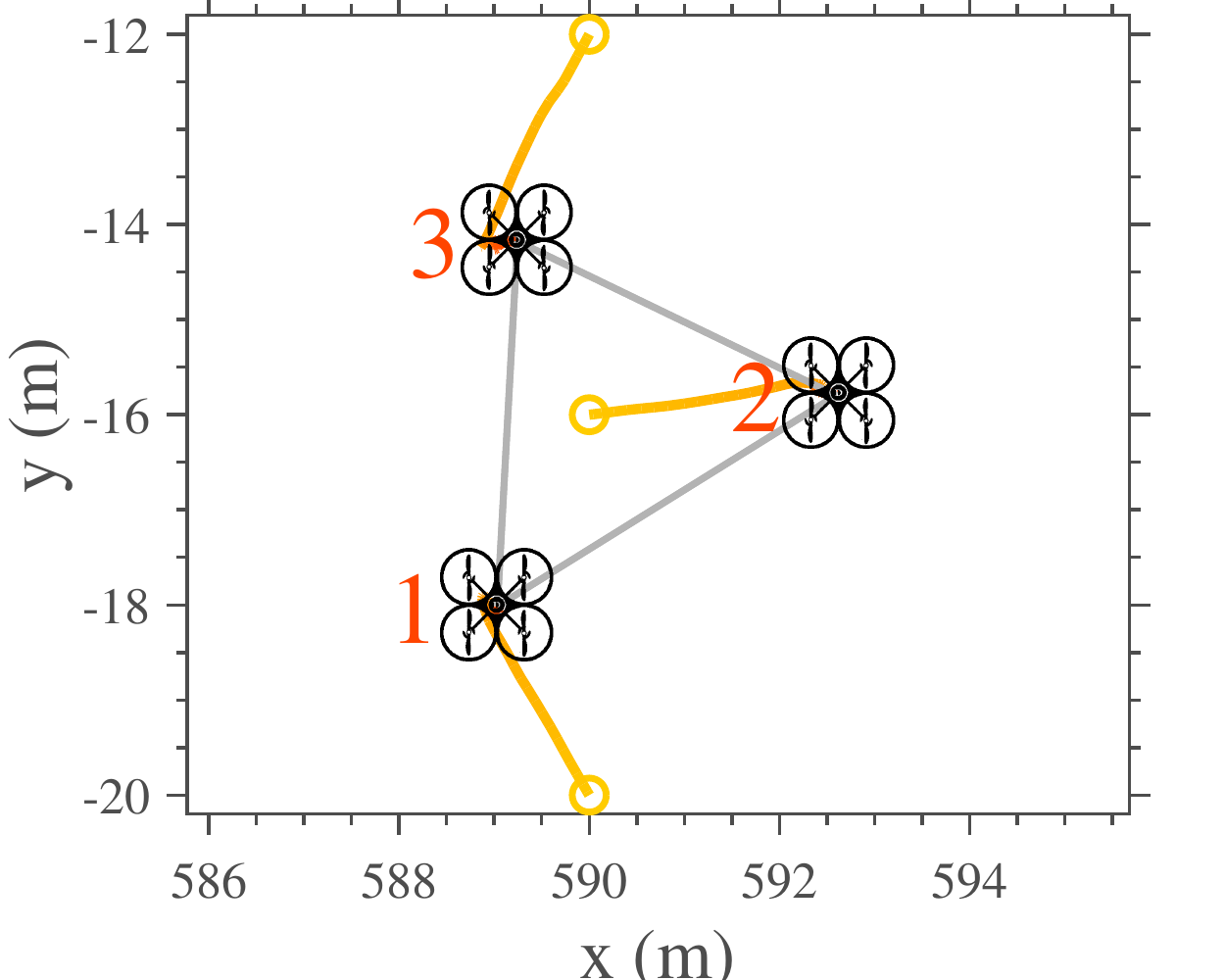}
		\caption{}
	\end{subfigure}%
	\begin{subfigure}[b]{0.5\linewidth}
		\includegraphics[trim = 1mm 1mm 1mm 1mm, clip, width=0.97\textwidth] {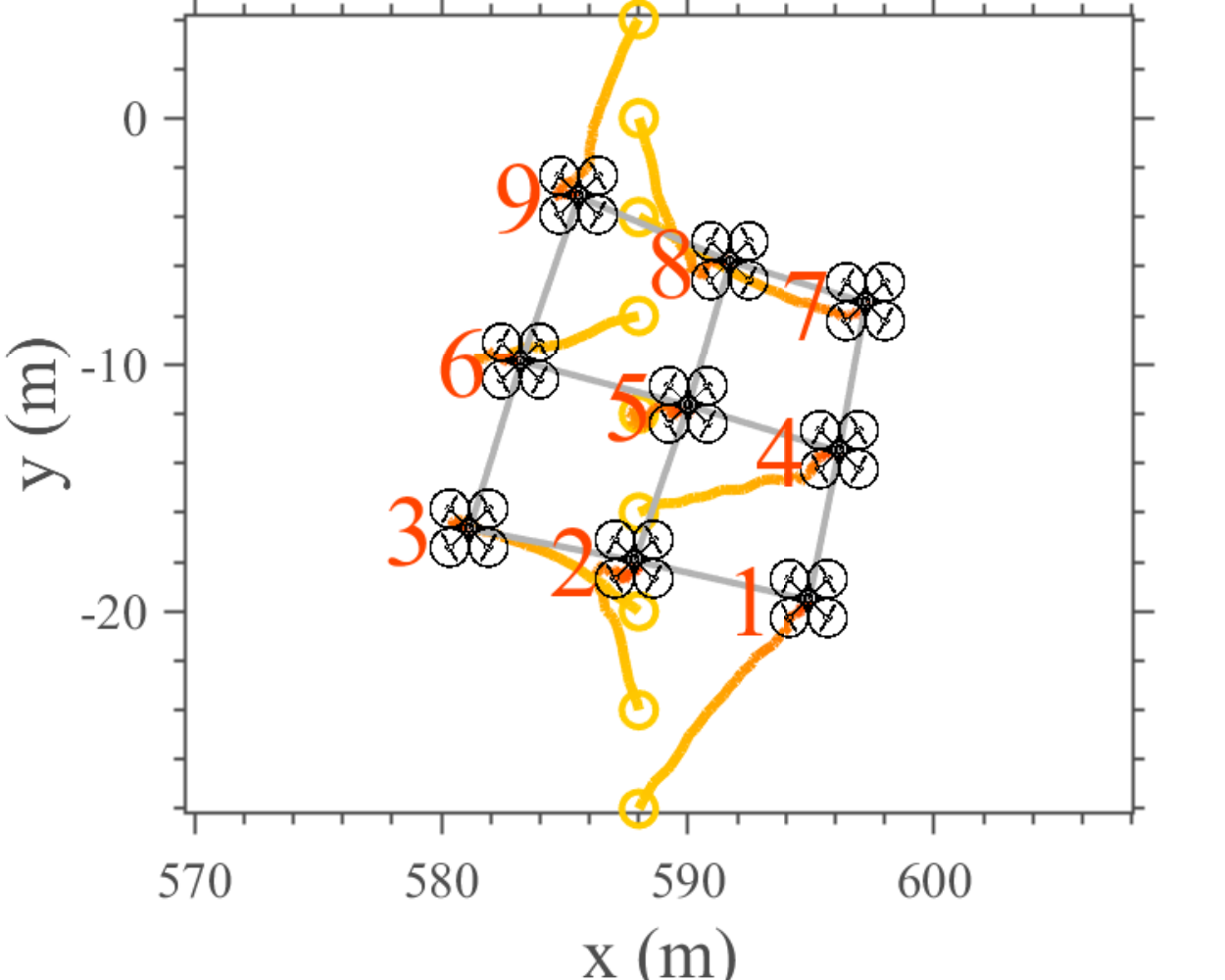}
		\caption{}
	\end{subfigure}%
	\caption{Trajectory of quadrotors under the proposed formation control pipeline.}
	\label{fig:Traj}	
\end{figure}

Fig.~\ref{fig:Sims}(a) shows a snapshot of the simulation for 3 quadrotors after reaching a desired equilateral triangle formation. The quadrotors initially start from a line configuration on the ground and ascend 20 meters. Only the downward-facing quadrotor cameras are used in this simulation to achieve the desired formation. Images captured by these cameras are shown in the left side of the figure, and an example of matched image feature point for the first and second quadrotors is shown in the bottom.  The ground truth trajectory of quadrotors based on the GPS data provided by the simulation software is shown in Fig.~\ref{fig:Traj}(a), where the communication topology among UAVs is shown via gray lines. Note that global position or orientation measurements (such as GPS or magnetometer) are not used in the formation control pipeline and are recorded only for plotting the  trajectories.

Fig.~\ref{fig:Sims}(b) shows a snapshot of the simulation for 9 quadrotors after reaching a desired square grid desired formation. In this simulation, the quadrotors ascend 40 meters from an initial line configuration, and only the front-facing cameras are used. Images captured by these cameras are shown in the left side of the figure, and the GPS trajectory of quadrotors is shown in Fig.~\ref{fig:Traj}(b).

Although the relative position vectors that are recovered up to an unknown positive scale factor from the QuEst algorithm are sufficient to achieve a desired formation, the collision avoidance strategy requires an estimate of the distance between two UAVs to detect an imminent collision. The altitude of the UAVs from the ground are used in our simulations to estimate the unknown scale factor for the recovered positions. In practice, onboard sensors such as altimeter, sonar, stereo cameras, laser range finder, etc., can be used to estimate the altitude.

\section{Future Work} \label{sec:Conclusion}

Our demonstrations in this work were focused on planar formations. We are currently working on extending the proposed pipeline to 3D formations. Our initial results have validated that our design approach can be used to achieve a desired 3D formation while avoiding collisions. 
Due to the distributed nature of our collision avoidance strategy gridlocks can occur. Incorporating a communication scheme such that agents can detect/avoid the gridlock situations will be a topic of future work.
Further future research include leveraging leader-follower strategies for cooperative navigation of multiple UAVs, hence, allowing a single human operator to navigate a team of UAVs while they autonomously travel in a specified formation.

\bibliographystyle{IEEEtran}
\bibliography{msBibs}

\end{document}